\documentclass[review]{elsarticle}

\usepackage{hyperref}
\usepackage{bigstrut,multirow,rotating,color}
\usepackage{geometry}
\usepackage{subfigure}
\usepackage{setspace}
\usepackage{amsmath}
\usepackage{ulem}

\journal{Journal of \LaTeX\ Templates}

\begin{document}
	
	\begin{frontmatter}
		
		\title{One-Two-One Networks for Compression Artifacts Reduction in Remote Sensing}
		
		
		\author[mymainaddress]{Baochang Zhang }
		\author[mymainaddress]{Jiaxin Gu }
		\author[myfourthaddress]{Chen Chen}
		\author[myfifthaddress]{Jungong Han}
		\author[mymainaddress]{Xiangbo Su }
		%
		
		\author[mysecondaryaddress]{Xianbin Cao }

		\author[mysixthaddress]{Jianzhuang Liu}
		
		\address[mymainaddress]{School of Automation Science and Electrical Engineering\\
			Beihang University, Beijing, China}
		\address[mysecondaryaddress]{School of Electronics and Information Engineering\\
			Beihang University, Beijing, China}
		\address[myfourthaddress]{Center for Research in Computer Vision (CRCV),
			University of Central Florida, Orlando, FL, USA}
		\address[myfifthaddress]{School of Computing and Communications, Lancaster University, LA1 4YW, UK.}
		\address[mysixthaddress]{Noah's Ark Lab, Huawei Technologies Co. Ltd., China}

		\begin{abstract}
			Compression artifacts reduction (CAR) is a challenging problem in the field of remote sensing. Most recent deep learning based methods have demonstrated superior performance over the previous hand-crafted methods. In this paper, we propose an end-to-end one-two-one (OTO) network, to combine different deep models, i.e., summation and difference models, to solve the CAR problem. Particularly, the difference model motivated by the Laplacian pyramid is designed to obtain the high frequency information, while the summation model aggregates the low frequency information. We provide an in-depth investigation into our OTO architecture based on the Taylor expansion, which shows that these two kinds of information can be fused in a nonlinear scheme to gain more capacity of handling complicated image compression artifacts, especially the blocking effect in compression. Extensive experiments are conducted to demonstrate the superior performance of the OTO networks, as compared to the state-of-the-arts on remote sensing datasets and other benchmark datasets. The source code will be available here\footnote{https://github.com/bczhangbczhang/}.
		\end{abstract}
		
		
	\end{frontmatter}
	
	
	\section{Introduction}\label{Introduction}
	In remote sensing, the satellite- or aircraft-based sensor technologies are used to capture and detect objects on Earth. Thanks to various propagated signals (e.g.,  electromagnetic radiation), remote sensing makes the data collection from dangerous or inaccessible areas possible, and therefore plays a significant role in many applications including monitoring, military information collection and land-use classification  \cite{Chen2016sivp,li2014novel,Li2015Local,vosselman2017contextual}. With the technological development of various satellite sensors, the volume of high-resolution remote sensing image data is increasing rapidly. Hence, proper compression of the satellite image becomes essential, which enables information exchange much more efficient, given a limited band width. 
	
	Existing compression methods generally fall into two categories: lossless (e.g., PNG) and lossy (e.g., JPEG) \cite{wang2002image}. The lossless methods usually provide better visual experience to users, but lossy methods often achieve higher compression ratios via non-invertible compression functions along with trade-off parameters to balance the data amount and the decompressed quality. Therefore the lossy compression schemes are always preferred by consumer devices in practice due to higher compression rate \cite{wang2002image}. However, high compression rate comes with the cost of having compression artifacts on the decoded image, which is a barrier for many applications, such as image analysis. Therefore, there is a clear need for compression artifact reduction, which is able to gain visual quality of the decompressed image, which can influence the visual effect and low-level vision processing \cite{yu2016deep}.
	
	\begin{figure}[h!]
		\begin{center}
			\includegraphics[width=0.7\linewidth]{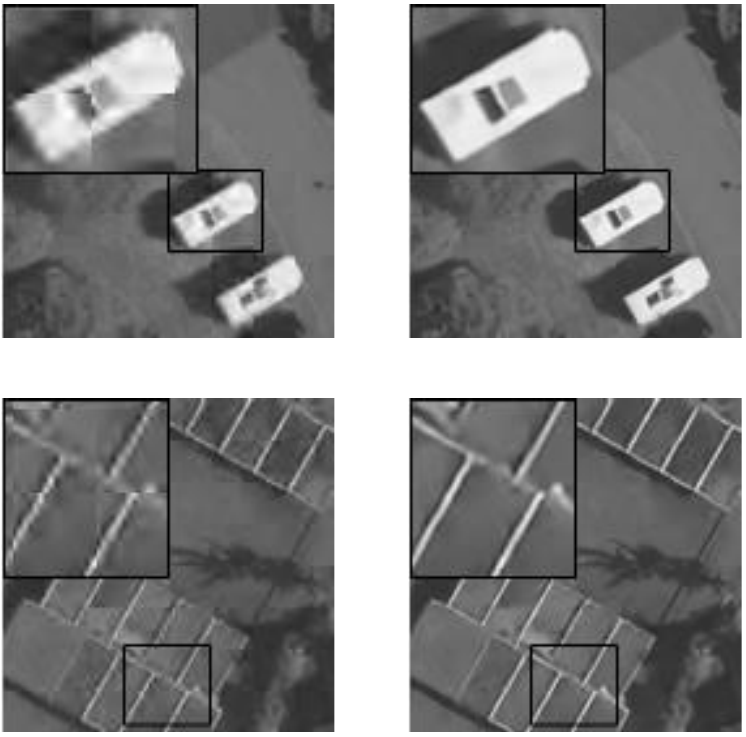}
			\caption{Left: the SPIHT-compressed remotely sensed images with obvious blocking artifacts. Right: the restored images by our OTO network, where lines are sharp and blurring is removed.}
			\label{fig:fig1}
		\end{center}
	\end{figure}
	
	The compression artifacts are in relation to the schemes used for compression. Take JPEG compression as an example, blocking artifacts are caused by discontinuities at the borders when encoding adjacent $8\times8$ pixel blocks, which are in the form of ringing effects and blurring due to the coarse quantization of the high frequency components. To deal with these compression artifacts, an improved version of JPEG, named JPEG 2000, is proposed, which adopts the wavelet transform to avoid blocking artifacts, but still undergoes ringing effects and blurring. 
	As an excellent alternative, SPIHT \cite{said1996new} showed that using simple uniform scalar quantization, rather than complicated vector quantization, also yields superior results. Due to its simplicity, SPIHT has been successful on natural (portraits, landscape, weddings, etc.) and medical (X-ray, CT, etc.) images. Furthermore, its embedded encoding process has proved to be effective in a broad range of reconstruction qualities. For instance, it can code fair-quality portraits and high-quality medical images equally well (as compared with other methods in the same conditions). However, in the field of remote sensing, the images usually suffer from severe artifacts after compression as shown in Fig.~\ref{fig:fig1}, which poses challenges to many high-level vision tasks, such as object detection \cite{Cheng201611,xiao2016change}, classification \cite{Chen2016sivp,Bian2017}, and anomaly detection \cite{Chein2002}.
	
	To cope with various compression artifacts, many conventional approaches have been proposed, such as filtering approaches  \cite{list2003adaptive},  \cite{reeve1984reduction},  \cite{wang2013adaptive}, specific priors (e.g., the quantization table in DSC \cite{liu2015data}), and thresholding techniques  \cite{liew2004blocking,foi2007pointwise}. Inspired by the great success of deep learning technology in many image processing applications, researchers start to exploit this powerful tool to reduce the compression artifact. Specifically, the Super-Resolution Convolutional Neural Network (SRCNN) \cite{dong2014learning} exhibits great potential of an end-to-end learning in image super-resolution. It is also pointed out that conventional sparse-coding-based image restoration model can be equally seen as a deep model. However, if we directly apply SRCNN to the compression artifact reduction task, the features extracted by its first layer are noisy, which will cause undesirable noisy patterns in reconstruction. Thus the three-layer SRCNN is not suitable for compressed image restoration, especially when dealing with complex artifacts. Thanks to transfer learning, ARCNN  \cite{yu2016deep} has been successfully applied to image restoration tasks. However, without exploiting the multi-scale information, ARCNNs fail to solve more complicated compression artifact problems.  Although many deep models with different architectures have been explored (e.g., \cite{dong2014learning,yu2016deep,cavigelli2017cas}) to solve the artifact reduction problem, there is little work incorporating different models in a unified framework to inherit their respective advantages.
	
\begin{figure*}[h!]
		\begin{center}
			\includegraphics[width=0.95\linewidth]{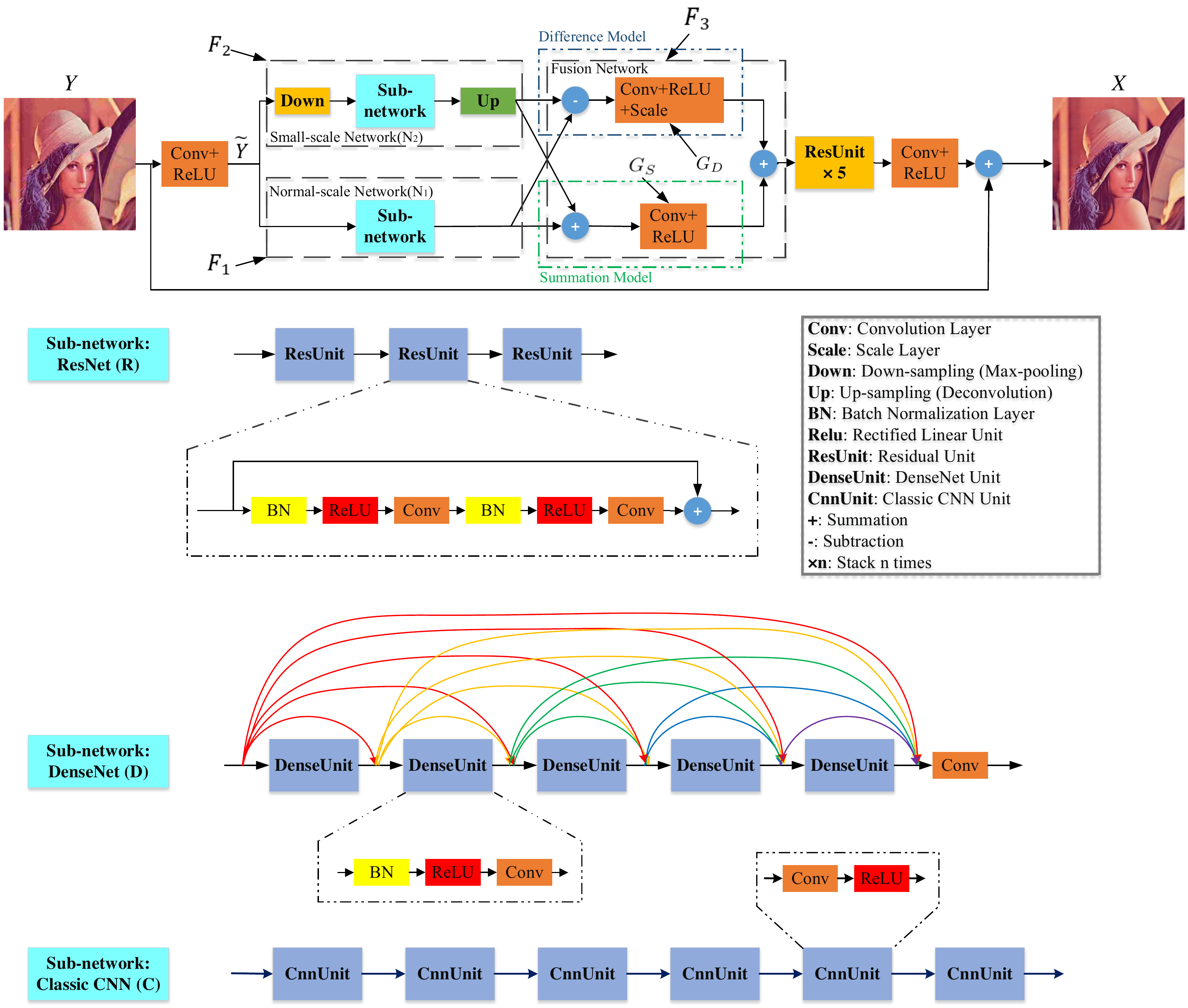}
			\caption{The architecture of One-Two-One (OTO) Network. Two different CNN models are combined in a principled framework, the outputs of which are further processed based on a fusion network. The details of the three sub-networks are also included. }
			\label{fig:OTO network}
		\end{center}
	\end{figure*}

	
	In this paper, a generic fusion network, dubbed as one-two-one (OTO) network, is developed for complex compression artifacts reduction. The general framework of the proposed OTO network is presented in Fig.~\ref{fig:OTO network}. Specifically, it consists of three sub-networks: a normal-scale network, a small-scale network with max pooling to increase the network receptive field, and a fusion network to perform principled fusion of the outputs from the summation and difference models. The summation model aggregates the low frequency information captured from different network scales, while the difference model is motivated by the Laplacian pyramid which is able to describe the high frequency information, such as detailed information. By combining the summation and difference models, both low and high frequency information of the image can be better characterized. This is motivated by the fact that adopting different schemes to process high frequency and low frequency information always benefits to low-level image processing applications, such as image denoising \cite{Zhang2017Beyond} and image reconstruction \cite{early2001image}. Most importantly, we provide an in-depth investigation into our OTO architecture based on the Taylor expansion, which shows that these two kinds of information are fused in a nonlinear scheme to gain more capacity to handle complicated image compression artifacts. From a theoretical perspective, this paper proposes a principled combination of different CNN models, providing the capability of coping with the extremely challenging task of the large blocking effect. Extensive experimental results verify that combining diverse models effectively boosts the performance. In a summary, we have the following contributions in this paper.

	\begin{enumerate}
		\item We develop a new  one-two-one (OTO) network, to combine different models based on an end-to-end deep framework aiming to effectively deal with complicated artifacts, i.e., the big blocking effect in compression.
		\item We are motivated by the idea of the Laplacian pyramid, which is extended in the deep learning framework, and explicitly used to capture the high frequency information in images. We show in the experiments that the difference model is able to effectively improve the compression artifact reduction performance.
		\item Based on the Taylor expansion, we lead to two OTO variants, which provide a profound investigation into our method. 
		\item Extensive experiments are conducted to validate the performance of OTO over the state-of-the-arts on both the benchmark datasets and remote sensing datasets.
	\end{enumerate}
	
	\begin{table*}
		\caption{A brief description of variables used in the paper.}
		\centering
		\begin{tabular}{|ll|}
			\hline
			$Y$: input compressed image   & $F_1$: normal-scale network \\
			$\tilde{Y}$: output of the first convolutional layer  &  $F_2$: small-scale network\\
			$N_1$: output of $F_1$ &   $N_2$: output of $F_2$  \\
			$Sum$: summation of $N_1$ and $N_2$  &  $Dif$: difference between $N_1$ and $N_2$  \\
			$G_S$: nonlinear operation of the summation model & $G_D$: nonlinear operation of the difference model\\
			$H_S$: output of $G_S$ &   $H_D$: output of $G_D$  \\
			$F_3$: fusion network &  $\alpha$: weight term between the two sub-networks\\
			$G_S'(\cdot)$: derivative of $G_S$  &  $G_D'(\cdot)$: derivative of $G_D$\\
			$\gamma$: constant term  &  $o[\cdot,\cdot]$: higher order infinitesimal\\
			$X$: uncompressed target image  &  \\
			\hline
		\end{tabular}
		\label{tab:variables}
	\end{table*}
	
	For ease of explanation, we summarize main variables in Table \ref{tab:variables}. The rest of the paper is organized as follows. Section \ref{Related work} introduces the related works, and section \ref{One-Two-One Networks} describes the details of the proposed method. Experiments and results are presented in section \ref{Implementation and Experimentsks}. Finally, section \ref{Conclusion and future work} concludes the paper.

	\section{Related work}\label{Related work}
	The OTO network is proposed to combine  summation and difference models in the end-to-end framework. Particularly, the difference model motivated by the Laplacian pyramid is designed to obtain the high frequency information, while the summation model aggregates the low frequency information. Compared to the summation model, the difference model can provide more detailed information. In this section, we briefly described the related work about how the high frequency information used in the low-level image processing, and also the previous CAR methods.
	
	  \textbf{On the high frequency information.} The high frequency information has been exploited in tasks such as pansharping \cite{Vivone2015A}, superresolution \cite{kim2016accurate} and denoising \cite{Wang2017Dilated}. However, the way of exploring it is different from ours. Specifically, in image superresolution, a low resolution input image is first interpolated to have the same size of the high resolution image as input. Then the goal of the network becomes learning the high resolution image from the interpolated low resolution image \cite{kim2016accurate}. In other words, the network essentially aims to learn the high frequency information in order to obtain the high resolution output \cite{Lim2017Enhanced,Ledig2016Photo}. In pansharpening, the high frequency details are not available for multispectral bands, and must be inferred through the model \cite{Masi2017CNN,Scarpa2017Target,Vivone2015A} starting from those of Pan images. In denoising, residual learning is utilized to speed up the training process as well as boost the denoising performance \cite{Zhang2017FFDNet,Zhang2017Beyond,Wang2017Dilated}.
	 The Laplacian pyramid is ubiquitous for decomposing images into multiple scales and is widely used for image analysis \cite{paris2011local, burt1983laplacian},
	which is computed as the difference between the original image and the low pass filtered image. This process is continued to obtain a set of band-pass filtered images, since each is the difference between two levels of the Gaussian pyramid. The Laplacian pyramids have been used to analyze images at multiple scales for a broad range of applications such as compression \cite{burt1983laplacian}, texture synthesis \cite{heeger1995pyramid}, and harmonization~\cite{sunkavalli2010multi}.
	
	
	\textbf{Traditional CAR methods.} Traditional methods for the CAR problem are generally categorized into deblocking-based and dictionary-based algorithms. The deblocking-based algorithms mainly focus on removing blocking and ringing artifacts using filters in the spatial domain or utilizing wavelet transforms and setting thresholds at different wavelet scales in the frequency domain. Among them, the most successful work is Shape-Adaptive Discrete Cosine Transformation (SA-DCT)  \cite{foi2007pointwise}, which achieved the state-of-the-art performance during the 2000s. However, similar to other deblocking-based methods, SA-DCT suffers from blurry edges and smooth texture regions as well. It is worth noting that SA-DCT is an unsupervised method, which is more powerful than supervised methods when there are not enough samples available. The supervised dictionary-based algorithms, such as RTF  \cite{jancsary2012loss}, S-D2  \cite{liu2015data}, take compression artifacts reduction as a restoration problem and reverse the impact of DCT-domain quantization by learned dictionaries. Unfortunately, the optimization procedure of sparse-coding-based approaches is always complicated and the end-to-end training does not seem to be possible, which limits their reconstruction performance.
	
	\textbf{Deep CAR methods.} Recently, deep convolutional neural networks have shown promising performance on both high-level vision tasks, such as classification  \cite{krizhevsky2009learning,cheng2017remote}, detection  \cite{ren2015faster,yao2015coarse,han2015object,cheng2016learning} and segmentation  \cite{long2015fully,yao2017revisiting,yao2016semantic}, and low-level image processing like super-resolution \cite{kim2016accurate}. Super-Resolution Convolutional Neural Networks (SRCNN) \cite{dong2014learning} utilize a three-layer CNN to increase the resolution of images and achieve superior results over the traditional SR algorithms like A+  \cite{timofte2014a+}. Following the idea of SRCNN, Yu et al. \cite{yu2016deep} eliminate the undesired noisy patterns by directly applying SRCNN architecture for compression artifacts suppression and prove that transfer learning also succeeds in low-level vision problems. Compression artifacts reduction CNN \cite{yu2016deep} mainly benefits from transfer learning in three aspects: from shallow networks to deep networks, from high-quality training datasets to low-quality ones and from one compression scheme to another scheme. Svoboda et al. \cite{svoboda2016compression} learn a feed-forward CNN by combining residual learning, skip architecture and symmetric weight initialization to improve image restoration performance. 
	The generative adversarial network (GAN) is also successfully used to solve the CAR problem. In \cite{gan2017}, the Structural Similarity (SSIM) loss is devised, which is a better loss with respect to the simpler Mean Squared Error (MSE), to re-formulate the compression artifact removal problem in a generative adversarial framework. The method obtains better performance than MSE trained networks.
	
	Due to the fixed quantization table in the JPEG compression standard, it is reasonable to take advantage of JPEG-related prior for better restoration performance. Deep Dual-domain Convolutional neural Network (DDCN)  \cite{guo2016building} adds DCT-domain prior into the dual networks so that the network is able to learn the difference between the original images and compressed images in both pixel-domain and DCT-domain. Likewise, D3 method \cite{wang2016d3} converts sparse-coding approaches into an LISTA-based  \cite{gregor2010learning} deep neural network, and gains both speed and performance. Both of DDCN and D3 adopt JPEG-related priors to improve reconstruction quality.
	One-to-many network \cite{one2many2017} is proposed for compression artifacts reduction. The network consists of three losses, a perceptual loss, a naturalness loss, and a JPEG loss, to measure the output quality. By combining multiple different losses, the one-to-many network is able to achieve visually pleasing artifacts reduction.
	
	\textbf{Challenges of the CAR problem.} In spite of already achieving good compression artifact removal performance, they still have limitations, especially when dealing with satellite imagery. Prior-based methods may not be generalized to other compression schemes like SPIHT, and therefore their applications are limited for the reason that satellite- or aircraft-based sensor technologies use variable compression standards. Another ignored problem-specific prior is the size of blocks, which is typically $8\times8$. The existing JPEG-based methods crop images into sub-samples or patches with small size like $32\times32$ and use $8\times8$ blocks for processing. However, larger block size like $32\times32$ is often adopted in the digital signal processor (DSP) of satellites for parallel processing. In this case, an image patch only contains a whole block and might have negative impact on the training process. As a result, it is important for sub-samples to contain several blocks so that the networks can perceive the spatial context between adjacent blocks.
	On the other hand, the existing deep learning based compression artifact removal approaches mainly focus on the architecture design \cite{yu2016deep,dong2014learning,svoboda2016compression} or changing the loss function \cite{gan2017,one2many2017}, with no theoretical explanations so that they fail to provide more profound investigation into methodologies. Moreover, the benefits of different network architectures are not fully explored for solving the CAR problem. 

	
	\section{One-Two-One Networks}\label{One-Two-One Networks}
	
	The OTO networks are designed to reduce compression artifacts based on a unified framework. As shown in Fig.~\ref{fig:OTO network}, two different models (summation model and difference model) are used to restore the input image individually, whose advantages are inherited by a CNN fusion network, and thus leading to a better performance than using each of them individually. In what follows, we address two issues to build the OTO network. We first describe the motivation of OTO, along with a theoretical investigation into the network architecture which leads to two variants. We then elaborate the architectures of the proposed OTO network, which are divided into three specific sub-networks. For each of them, we give the details of the implementation.

	\subsection{Theoretical Investigation of OTO}\label{Theoretical Investigation of OTO}
	
	OTO is a general framework aiming to combine different deep models of different architectures. In OTO, a hierarchical CNN structure is exploited to capture multi-scale texture information, which is very effective in dealing with various compression artifacts. In addition, each network in our framework carries out a specific objective, i.e., different-scale textures, and we end up combining them together to obtain better results. 
	The idea origins from the Laplacian pyramid for capturing detailed information, but we use the different scale networks to implement the idea in the deep learning framework. The small-scale network involves spatial max pooling, which essentially increases the network receptive field and aggregates information in larger spatial area. Therefore, by combining small-scale network and normal-scale network features, the network learns features from different scales. 
	Inspired by the Laplacian pyramid, the difference model is exploited in the deep framework and able to describe the high frequency information, while the summation model captures the low frequency information. We then combine both in a principled end-to-end deep framework. We like to highlight our idea from a more basic way. We provide a sensible way to combine the low and high frequency information in the deep learning framework, and also theoretically explain it with the Taylor expansion. In OTO, we have:
	\begin{equation}
	N_1=F_1\left(\tilde{Y}\right),
	\label{eq:1}
	\end{equation}
	and
	\begin{equation}
	N_2=F_2\left(\tilde{Y}\right),
	\label{eq:2}
	\end{equation}
	where $\tilde{Y}$ is the output of the first convolutional network, which is designed to pre-process the input compressed image $Y$ based on convolution layers. $N_1$ and $N_2$ denote the outputs of the two branch networks, i.e., normal-scale network and small-scale network.  To better restore the input image $X$, we exploit two different networks, i.e., summation model and difference model, based on $\tilde{Y}$, which complement each other in terms of different network architectures.  The summation model is used to mitigate the disparity between two networks, while the difference model highlights that different CNNs are designed for different purposes in order to obtain better restoration results. We have:
	\begin{equation}
	Sum=N_1+N_2,
	\label{eq:3}	
	\end{equation}
	which actually aggregates the low frequency information. 
	\begin{equation}
	Dif=N_1-N_2,
	\label{eq:4}	
	\end{equation}
	which describes the high frequency information as shown in the Laplacian pyramid. $G_S$ and $G_D$ denote the two branches following the summation and subtraction operation in Fig.~\ref{fig:OTO network} respectively. Both kinds of information are then combined together for a better restoration performance, and we have:
	\begin{equation}
	H_S=G_S\left(Sum\right),
	\label{eq:5}	
	\end{equation}
	and
	\begin{equation}
	H_D=G_D\left(Dif\right),
	\label{eq:6}	
	\end{equation}
	where $H_S$ and $H_D$ are the outputs of the two branches. They are then combined together via a nonlinear operation, which is designed to be robust to the artifacts in the compressed images. And we have:  
	
	\begin{equation}
	\begin{aligned}
	F_3\left(\tilde{Y}\right)=&H_S+\alpha H_D=G_S\left(Sum\right)+\alpha G_D\left(Dif\right)\\
	&=G_S\left(N_1+N_2\right)+\alpha G_D\left(N_1-N_2\right)
	\label{eq:7}
	\end{aligned}
	\end{equation}
	where $\alpha$ is a weight factor to balance different models. Based on Taylor expansion on $G_S$ and $G_D$, we prove that our OTO is actually the combination of $N_1$ and $N_2$ based on a nonlinear scheme as:
	\begin{equation}
	\begin{aligned}
	F_3\left(\tilde{Y}\right)= &G_S'\left(\left(N_1+N_2\right)^*\right)\times\left(N_1+N_2\right)+\alpha G_D'\left(\left(N_1-N_2\right)^*\right)\times\left(N_1-N_2\right)\\
	& +\gamma +o\left[\left(N_1+N_2\right),\left(N_1-N_2\right)\right]\\
	&=\left(G_S'\left(\left(N_1+N_2\right)^*\right)+\alpha  G_D'\left(\left(N_1-N_2\right)^*\right)\right)N_1 +\\
	& \left(G_S'\left(\left(N_1+N_2\right)^*\right)- \alpha G_D'\left(\left(N_1-N_2\right)^*\right)\right)N_2+\gamma+o\left[\left(N_1+N_2\right),\left(N_1-N_2\right)\right],
	\label{eq:8}
	\end{aligned}
	\end{equation}
	where $*$ means that there is a point, which is always differential, used in the Taylor expansion. 
		$\gamma$ denotes the constant term, and $o[(N_1+N_2 ),(N_1-N_2 )]$ denotes the higher order infinitesimal. More specifically,  $o[(N_1+N_2 ),(N_1-N_2 )]$  in Eq.~\ref{eq:8} is the nonlinear part and the remaining is the linear part. Note that the adopted nonlinear OTO model includes both  the linear  and nonlinear parts. 
	
\begin{figure*}
		\begin{minipage}[t]{.4\linewidth}
			\centering
			\includegraphics[width=7.8cm]{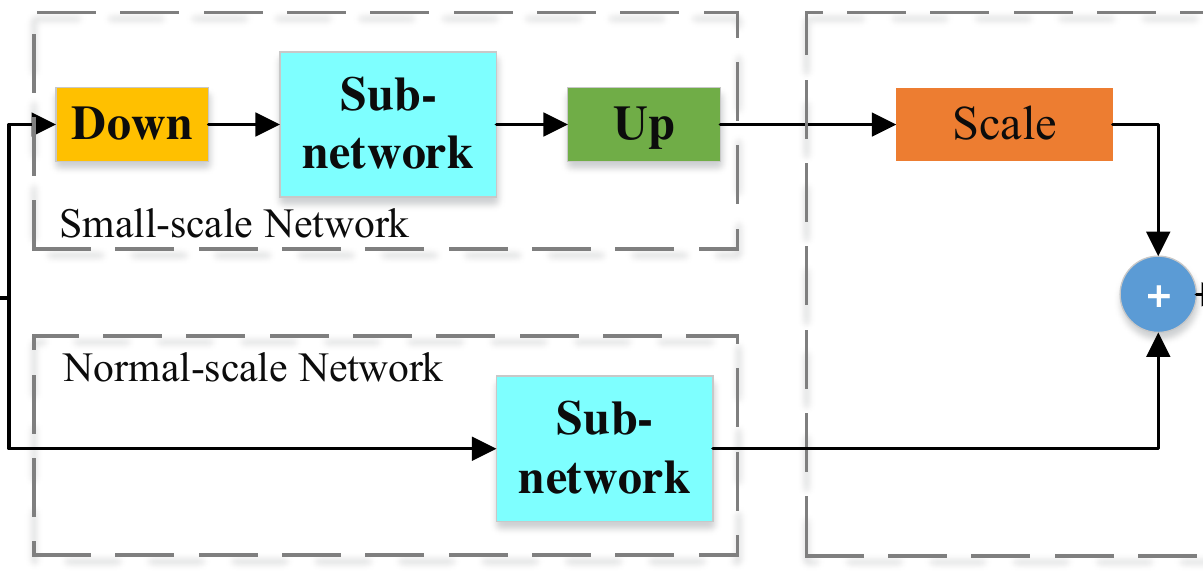}
			\caption{The architecture of OTO(Linear) with a learned $\alpha$ to balance the two branch networks.}
			\label{fig:linear}
		\end{minipage}
		\hfill
		\begin{minipage}[t]{.39\linewidth}
			\centering
			\includegraphics[width=6.1cm]{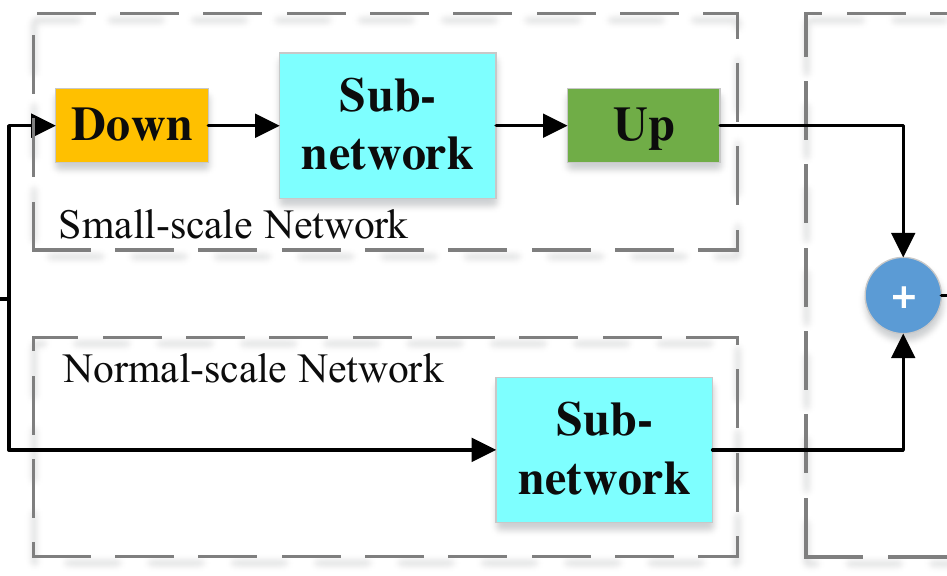}
			\caption{The architecture of OTO(Sum)  without the difference model.}
			\label{fig:sum}
		\end{minipage}
	\end{figure*}

	Based on Eq.~\ref{eq:8}, two linear OTO variants can be obtained as shown  in Fig.~\ref{fig:linear} and Fig.~\ref{fig:sum}. The first one, termed as OTO(Linear), is:
	\begin{equation}
	F_{3,1}\left(\tilde{Y}\right)=N_1+\alpha N_2,
	\label{eq:9}
	\end{equation}
	which can be derived from the linear part of Eq.~\ref{eq:8}.  In its implementation we learn  $\alpha$ that is elaborated in the experimental part. Particularly $\alpha=1$, we obtain the second one:
	\begin{equation}
	F_{3,2}\left(\tilde{Y}\right)=N_1+N_2,
	\label{eq:10}
	\end{equation}
	which leads to our baseline, termed as OTO(Sum). 
	
	\subsection{The architectures of OTOs}\label{The architectures of OTOs}
	
	There are three distinct parts in OTOs: a) normal-scale restoration network, b) small-scale restoration network, c) fusion network. For sub-networks a) and b), three kinds of CNN models are available: R, D and C (short for ResNet, DenseNet and Classic CNNs respectively). The details about the OTO network are shown in Fig.~\ref{fig:OTO network}.

	\textbf{ResNet(R):} For each ResUnit, we follow the latest variant proposed in  \cite{he2016identity}, which is more powerful than its predecessors. More specifically, in each ResUnit, batch normalization layer  \cite{ioffe2015batch}, ReLU layer  \cite{krizhevsky2012imagenet} and convolution layer are stacked twice in sequence. 

	\textbf{DenseNet(D):} Inspired by Densely Connected Convolutional Networks \cite{huang2016densely}, to further improve the information flow between layers we propose a different connectivity pattern: we introduce direct connections from any layer to all subsequent layers. In DenseNet, the feature fusion method is converted from addition to concatenation compared with ResNet, resulting in wider feature maps. The growth rate $k$ is an important concept in DenseNet which means how fast the width of feature maps grows and in our implementation, we set $k$ to 8. For each DenseUnit, we also follow the pre-activation style unit as ResUnit except the number of convolutional layers is reduced to 1. As can be seen in Fig.~\ref{fig:OTO network}, five DenseUnits are stacked sequentially followed by a convolutional layer to reduce the width of feature map so that it can be fused with the other sub-network.

	\textbf{Classic CNNs(C):} The classic CNN models only take advantages of convolutional layers and activation layers. The CnnUnit consists of one convolutional layer and one ReLU layer, and 6 CnnUnits are stacked to form the Classic CNN sub-network.
	
	In the sub-network b), we utilize $2\times2$ max-pooling to decrease the size of feature map by half, which obtains the following benefits:  the computational cost is decreased to 1/4, and with more robust features extracted compared to the sub-network a), and thus enlarging the perceptional field.

	\textbf{Fusion Network:} Following Eq.\ref{eq:7}, we construct the fusion network. Convolutional layers with ReLUs serve as the nonlinear operation, and scale layers serve as the weight term. After fusion, we stack 5 more ResUnits to further restore the images.

	\textbf{OTO Naming Rules:} For convenience, we use abbreviations to represent the three kinds of sub-networks. The first and second abbreviations after OTO represent the normal-scale and small-scale sub-networks respectively. For example, OTO\_RD stands for an OTO network whose normal-scale sub-network is a ResNet and whose small-scale sub-network is a DenseNet.

\begin{figure*}[h!]
		\begin{center}
			\includegraphics[width=0.95\linewidth]{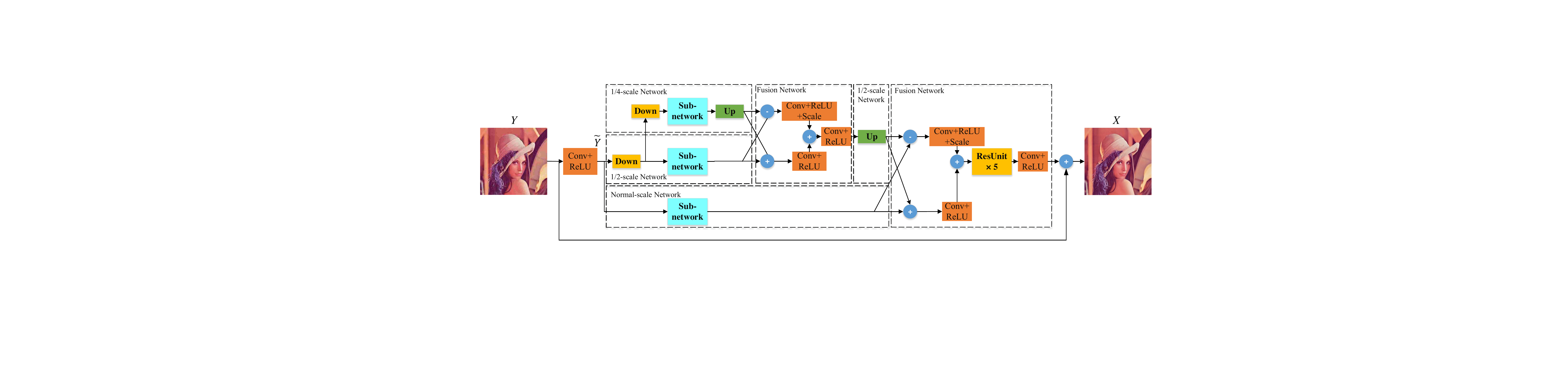}
			\caption{The architecture of multi-scale OTO network (OTO\_RRR) in which $\times1$,$\times \frac{1}{2}$ and $\times \frac{1}{4}-$scale features are exploited. The sub-networks are the same as in Fig.~\ref{fig:OTO network}.}
			\label{fig:OTO Multi-scale}
		\end{center}
	\end{figure*}

	\textbf{Multi-scale OTO Networks:} To further investigate our proposed OTO network, we design a multi-scale network whose structure is shown in Fig.~\ref{fig:OTO Multi-scale}.  $\times\frac{1}{2}$ and $\times\frac{1}{4}$-scale features are first fused by the first fusion network to get the combined $\times\frac{1}{2}$-scale feature. Then the fused feature along with the $\times1$-scale feature serves as the input of the second fusion network. Except for the architecture, all the other details are the same as the two-scale OTO network.
	
	It should be noted that the proposed OTO network exploits a series of ResUnits to fit the residual of the input and target images. In other words, there is a long and direct shortcut connecting the input image and the output of the subsequent network apart from the identity shortcut of each ResUnit. VDSR \cite{kim2016accurate} has already proved that learning the residual between low-resolution  and high-resolution image is more efficient and effective in the super-resolution task, because the difference is small, i.e., residual is sparse. In the CAR task, this intuition is tenable because the compression algorithms do not change the essence of the image. As a result, the ResUnit leads to a sparse residual and thus we can train the network efficiently.

	\section{Implementation and Experiments}\label{Implementation and Experimentsks}
	
	\subsection{Datasets}\label{Datasets}
	
	In order to evaluate the OTO network, three groups of training and test dataset settings are designed, which are given according to their different test sets.
	
	\textbf{LIVE1 and Classic 5:} Following the protocol of ARCNN \cite{yu2016deep}, we evaluate our proposed network based on BSD500 \cite{arbelaez2011contour}, where the training and test set are combined to form a 400-image training set and a 100-image validation set. The disjoint dataset LIVE1  \cite{sheikh2005live} containing 29 images is chosen as our test set. Another test set we adopt is Classic 5, one of the most frequently used datasets for evaluating the quality of images. 
	
	\textbf{BSD500 Test set:} The BSD500 test set contains 200 images, which is more challenging than LIVE1, and more widely adopted in the recent research papers  \cite{guo2016building}. Considering that the 200-image  BSD500 training set is too small to generate enough sub-samples when a large stride is chosen, we perform data augmentation by rotating the original images by 90, 180 and 270 degrees. The remaining 100-image BSD500 validation set is used for validation. 
	
	\textbf{Remotely Sensed Datasets:} There are two public remote sensing datasets on ``ISPRS Test Project on Urban Classification and 3D Building Reconstruction": ``Downtown Toronto" and ``Vaihingen"  \cite{cramer2010dgpf}. To validate the performance of OTO on remote sensing images, ``Downtown Toronto" dataset is employed, which contains various landscapes, such as ocean, road, house, vehicle and plant. To build a dataset for the CAR problem, we preprocess the high-resolution images in the ``Downtown Toronto" dataset by using various compression algorithms, but obviously without the need for labeling the ground truth. SPIHT compression algorithm is used. The SPIHT algorithm can be applied to satellite images, where the original images is cropped into sub-images with a specific size $32\times32$. Compared to JPEG, the size of block artifacts in SPIHT is $32\times32$, which is much larger than that used in JPEG. It is different from the quality factor in JPEG that the compression degree is decided by compression ratio, such as 8, 16, 32 and 64. Afterwards, we build the datasets used for training and validation. We randomly pick up 400 non-overlapping sub-images from the source images and the compressed images to form the training set and each image has a uniform size of $512\times512$. Then we do the same operation to get the 200-image disjoint validation set. For testing, we use the other dataset ``Vaihingen" to build a 400-image test set that has the same setting as the training set. 

	\textbf{Evaluation Metrics:} To quantitatively evaluate the proposed method, three widely used metrics: peak signal-to-noise ratio (PSNR), PSNR-B  \cite{yim2011quality} and structural similarity (SSIM)  \cite{wang2004image} are adopted in our experiments. PSNR is an engineering term for the ratio between the maximum possible power of a signal and the power of corrupting noise that affects the fidelity of its representation, which is most commonly used to measure the quality of reconstructed image after a lossy compression. The PSNR-B modifies PSNR by including a blocking effect factor resulting in a better metrics than PSNR for quality assessment of impaired images. SSIM index is a method for predicting the perceived quality of digital images. SSIM considers image degradation as perceived change in structural information. While incorporating important perceptual phenomena, it also includes both luminance masking and contrast masking terms.
	
	\textbf{Other Settings:} We only focus on restoring the luminance channel of the compressed image, and RGB-to-YCbCr operation is applied via MATLAB function. We also use MATLAB to carry out JPEG compression to generate compressed images with different qualities, such as QF-10, 20, 30 and 40. It is also worth noting that we crop every image such that the number of pixels in height and width are even since an odd number will affect the process of down-sampling and up-sampling (padding is necessary). 
	To train the proposed OTO network, we choose SGD as the optimization algorithm with a momentum 0.9 and a weight decay 0.001. The initial learning rate is 0.01 with a degradation of 10\% over every 30000 iterations before it reaches the maximum iteration number 120000.
	
	\subsection{Sub-networks and Multi-scale network}\label{Sub-networks and Multi-scale network}
	
	\begin{table}[htbp]
		\tiny
		\centering
		\caption{Results on different combination of sub-networks.  Red marks mean the best results and blue marks mean the second best results}
		\begin{tabular}{|c|c|c|c|c|c|c|c|c|c|}
			\hline
			LIVE1 & Quality & OTO\_CC & OTO\_CR & OTO\_RC & OTO\_DD & OTO\_DR & OTO\_RD & OTO\_RR & OTO\_RRR \bigstrut\\
			\hline
			\multirow{4}[8]{*}{PSNR} & 10    & 29.01  & 29.25  & 29.24  & 29.25  & 29.24  & 29.23  & \textcolor[rgb]{ 0,  .439,  .753}{29.28 } & \textcolor[rgb]{ 1,  0,  0}{\textbf{29.38 }} \bigstrut\\
			\cline{2-10}          & 20    & 31.39  & 31.61  & 31.61  & 31.63  & 31.63  & 31.61  & \textcolor[rgb]{ 0,  .439,  .753}{31.67 } & \textcolor[rgb]{ 1,  0,  0}{\textbf{31.79 }} \bigstrut\\
			\cline{2-10}          & 30    & 32.52  & 33.04  & 33.05  & 33.04  & 33.06  & 33.07  & \textcolor[rgb]{ 0,  .439,  .753}{33.08 } & \textcolor[rgb]{ 1,  0,  0}{\textbf{33.10 }} \bigstrut\\
			\cline{2-10}          & 40    & 33.56  & 34.07  & 34.08  & 34.09  & 34.09  & 34.09  & \textcolor[rgb]{ 0,  .439,  .753}{34.10 } & \textcolor[rgb]{ 1,  0,  0}{\textbf{34.13 }} \bigstrut\\
			\hline
			\multirow{4}[8]{*}{SSIM} & 10    & 0.8213  & 0.8289  & 0.8293  & 0.8293  & 0.8296  & 0.8293  & \textcolor[rgb]{ 0,  .439,  .753}{0.8298 } & \textcolor[rgb]{ 1,  0,  0}{\textbf{0.8301 }} \bigstrut\\
			\cline{2-10}          & 20    & 0.8851  & 0.8946  & 0.8949  & 0.8952  & 0.8951  & 0.8950  & \textcolor[rgb]{ 0,  .439,  .753}{0.8954 } & \textcolor[rgb]{ 1,  0,  0}{\textbf{0.8958 }} \bigstrut\\
			\cline{2-10}          & 30    & 0.9124  & 0.9212  & 0.9217  & 0.9215  & 0.9218  & 0.9218  & \textcolor[rgb]{ 1,  0,  0}{\textbf{0.9218 }} & \textcolor[rgb]{ 1,  0,  0}{\textbf{0.9218 }} \bigstrut\\
			\cline{2-10}          & 40    & 0.9274  & 0.9358  & 0.9361  & 0.9361  & \textcolor[rgb]{ 0,  .439,  .753}{0.9362 } & \textcolor[rgb]{ 0,  .439,  .753}{0.9362 } & \textcolor[rgb]{ 0,  .439,  .753}{0.9362 } & \textcolor[rgb]{ 1,  0,  0}{\textbf{0.9365 }} \bigstrut\\
			\hline
			\multirow{4}[8]{*}{PSNR-B} & 10    & 28.74  & 28.94  & 28.92  & 28.95  & 28.92  & 28.91  & \textcolor[rgb]{ 0,  .439,  .753}{28.95 } & \textcolor[rgb]{ 1,  0,  0}{\textbf{29.13 }} \bigstrut\\
			\cline{2-10}          & 20    & 31.03  & 31.09  & 31.14  & 31.15  & 31.14  & 31.14  & \textcolor[rgb]{ 0,  .439,  .753}{31.17 } & \textcolor[rgb]{ 1,  0,  0}{\textbf{31.29 }} \bigstrut\\
			\cline{2-10}          & 30    & 31.67  & 32.46  & 32.47  & 32.44  & 32.45  & 32.45  & \textcolor[rgb]{ 0,  .439,  .753}{32.48 } & \textcolor[rgb]{ 1,  0,  0}{\textbf{32.52 }} \bigstrut\\
			\cline{2-10}          & 40    & 33.04  & 33.40  & 33.43  & 33.44  & 33.44  & 33.45  & \textcolor[rgb]{ 0,  .439,  .753}{33.48 } & \textcolor[rgb]{ 1,  0,  0}{\textbf{33.53 }} \bigstrut\\
			\hline
		\end{tabular}%
		\label{tab:Sub-networks and Multi-scale network}%
	\end{table}%
	As mentioned before, the OTO network is a framework that can take advantage of any CNNs, e.g., ResNet(R), DenseNet(D) and Classic CNN(C), as its sub-networks. The results of combining different kinds of sub-networks are shown in Table \ref{tab:Sub-networks and Multi-scale network}. Classic CNNs obtain the worst results, but which can be improved by using different scales (OTO\_CC) or combining with {ResNet} (OTO\_CR and OTO\_RC). The OTO based on the densely connected network (OTO\_DD) is designed to encourage feature reuse, but the lack of an identity mapping  enforces the network to learn residual, which deems its failure. The combination of DenseNet and ResNet with different scales (OTO\_DR and OTO\_RD) are affected by two kinds of discriminated features. In contrast, residual learning benefits more on the CAR problem, and the combination of two ResNets (OTO\_RR) outperforms all other combinations. Multi-scale features show promising results, and we design a multi-scale OTO network (OTO\_RRR) by adding an 1/4-scale sub-network to OTO\_RR. The result outperforms OTO\_RR with a large margin on all three metrics. Even though OTO\_RRR has outstanding performance, its computational cost increases almost $25\%$,  resulting in more training and test time. After evaluating the pros and cons, we choose OTO\_RR as our main framework and if not mentioned, OTO means OTO\_RR in the following. We design an experiment by removing one of the sub-networks each time to investigate the function of the sub-networks. The results in Table \ref{tab:Ablation Experiment on Sub-networks} indicate that the normal-scale feature is shown to be more helpful than the small-scale feature when only one sub-network is adopted.
	\begin{table}[htbp]
		\centering
		\caption{Experiments on the sub-networks.}
		\begin{tabular}{|c|c|c|c|c|}
			\hline
			\textbf{LIVE1} & \textbf{Algorithm} & \textbf{PSNR} & \textbf{PSNR-B} & \textbf{SSIM} \bigstrut\\
			\hline
			\multirow{3}[6]{*}{\textbf{QF-10}} & \textbf{OTO(normal-scale)} & 28.65  & 27.91  & 0.8263  \bigstrut\\
			\cline{2-5}          & \textbf{OTO(small-scale)} & 28.26  & 27.62  & 0.8245  \bigstrut\\
			\cline{2-5}          & \textbf{OTO} & \textbf{29.28 } & \textbf{28.95 } & \textbf{0.8298 } \bigstrut\\
			\hline
		\end{tabular}%
		\label{tab:Ablation Experiment on Sub-networks}%
	\end{table}%

	\begin{figure}[h!]
		\begin{center}
			\includegraphics[width=0.95\linewidth]{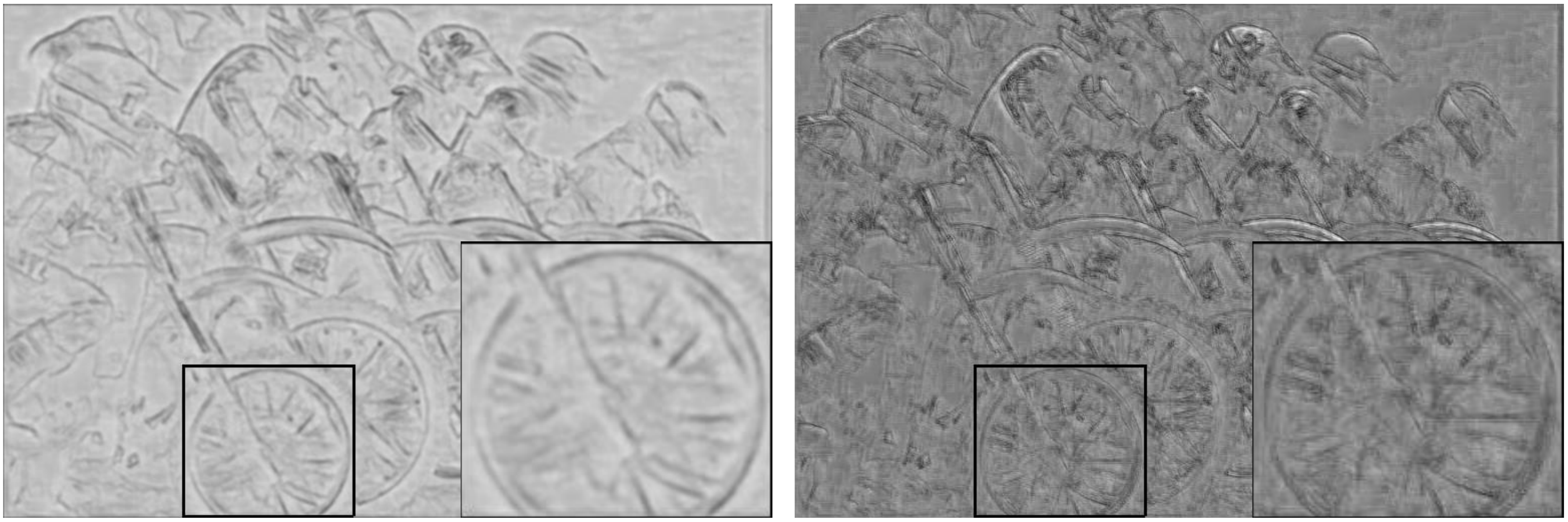}
			\caption{
			Left: the feature map of the summation model, which contains more low frequency information. Right: the feature map of the difference model, which provides more detailed information (high frequency). With the Fourier Transform, we can compare the amounts of the high frequency components between the two feature maps. By removing the DC (0 frequency) component from the frequency domain and considering those components with spatial frequencies $> 100$ as high frequency components (the size of the feature maps is 1067 $\times$ 1600), we can find that the ratio of the high frequency energy to the whole energy is about $ 38\% $ for the left map, while it is about $68\%$ for the right map.
			}
			\label{fig:frequency}
		\end{center}
	\end{figure}
	
	\subsection{OTO vs. its two Variants}\label{OTO Network vs. its two Variants}
	
	\begin{table}[htbp]
		\small
		\centering
		\caption{Comparative results between OTO and its two variants on LIVE1}
		\begin{tabular}{|c|c|c|c|c|}
			\hline
			\textbf{LIVE1} & \textbf{Algorithm} & \textbf{PSNR} & \textbf{PSNR-B} & \textbf{SSIM} \bigstrut\\
			\hline
			\multirow{4}[8]{*}{\textbf{QF-10}} & \textbf{JPEG} & 27.77  & 25.33  & 0.7905  \bigstrut\\
			\cline{2-5}          & \textbf{OTO(Linear)} & 29.26  & 28.91  & 0.8295  \bigstrut\\
			\cline{2-5}          & \textbf{OTO(Sum)} & 29.26  & 28.91  & 0.8287  \bigstrut\\
			\cline{2-5}          & \textbf{OTO} & \textbf{29.28 } & \textbf{28.95 } & \textbf{0.8298 } \bigstrut\\
			\hline
			\multirow{4}[8]{*}{\textbf{QF-20}} & \textbf{JPEG} & 30.07  & 27.57  & 0.8683  \bigstrut\\
			\cline{2-5}          & \textbf{OTO(Linear)} & 31.62  & 31.12  & 0.8950  \bigstrut\\
			\cline{2-5}          & \textbf{OTO(Sum)} & 31.60  & 31.07  & 0.8941  \bigstrut\\
			\cline{2-5}          & \textbf{OTO} & \textbf{31.67 } & \textbf{31.17 } & \textbf{0.8954 } \bigstrut\\
			\hline
			\multirow{4}[8]{*}{\textbf{QF-30}} & \textbf{JPEG} & 31.41  & 28.92  & 0.9000  \bigstrut\\
			\cline{2-5}          & \textbf{OTO(Linear)} & 33.06  & 32.45  & 0.9218  \bigstrut\\
			\cline{2-5}          & \textbf{OTO(Sum)} & 32.65  & 32.27  & 0.9160  \bigstrut\\
			\cline{2-5}          & \textbf{OTO} & \textbf{33.08 } & \textbf{32.48 } & \textbf{0.9218 } \bigstrut\\
			\hline
			\multirow{4}[8]{*}{\textbf{QF-40}} & \textbf{JPEG} & 32.35  & 29.96  & 0.9173  \bigstrut\\
			\cline{2-5}          & \textbf{OTO(Linear)} & 34.03  & 33.43  & 0.9349  \bigstrut\\
			\cline{2-5}          & \textbf{OTO(Sum)} & 33.68  & 33.35  & 0.9316  \bigstrut\\
			\cline{2-5}          & \textbf{OTO} & \textbf{34.10 } & \textbf{33.48 } & \textbf{0.9362 } \bigstrut\\
			\hline
		\end{tabular}%
		\label{tab:Comparisons between OTO and its two variants on LIVE1}%
	\end{table}%

	As mentioned above, OTO has the capability to utilize the nonlinear model, i.e., the summation and difference models, which is fully evaluated in this section. OTO(Linear) and OTO(Sum) are used in our comparison. The former one learns a weight factor to balance the significance of two branch networks, which adaptively combine two CNNs to solve the CAR problem. It is verified to be very effective for the reason that the significance of each CNN should be well considered in the fusion process. For the OTO(Sum) network, the weight factor is fixed to 1, which means that this version of OTO is not only shortage of the nonlinear representation ability but also impossible to tell which branch network contains more important information to suppress compression artifacts. In other words, OTO(Sum) just directly applies the addition operation to the two sub-networks. In this experiment, all comparative networks are trained based on BSD500 training and testing sets, and then tested on LIVE1 and Classic5. The results are shown in Table \ref{tab:Comparisons between OTO and its two variants on LIVE1} and Table \ref{tab:Comparisons between OTO and its two variants on Classic 5}. The OTO network along with its two variants have promising restoration performances on the four quality factors. These results demonstrate the effectiveness of the auto-learned weight factor $\alpha$ and the nonlinear operation on the summation and difference of the two branch networks. PSNR-B metric is designed specifically to measure the blocking artifacts. Particularly, we analyze the PSNR-B gain of OTO and OTO(Linear) compared to their baseline OTO(Sum).  We observe that for low-quality (QF-10, QF-20) compression images, the nonlinear operation benefit more than the weight factor on suppressing blocking artifacts, but different for high-quality images (QF-30, QF-40). We trained OTO models on GTX 1070, I7-6700k with 32G memory. The training time for OTO, OTO(Linear), and OTO(Sum) are 6h12m, 5h42m and 5h31m, respectively. The average test time for OTO, OTO(Linear), and OTO(Sum) are 0.1803s, 0.1738s and 0.1734s per image, respectively.
	\begin{table}[htbp]
		\small
		\centering
		\caption{Comparisons between OTO and its two variants on Classic 5}
		\begin{tabular}{|c|c|c|c|c|}
			\hline
			\textbf{Classic5} & \textbf{Algorithm} & \textbf{PSNR} & \textbf{PSNR-B} & \textbf{SSIM} \bigstrut\\
			\hline
			\multirow{4}[8]{*}{\textbf{ QF-10}} & \textbf{JPEG} & 27.82  & 25.21  & 0.7800  \bigstrut\\
			\cline{2-5}          & \textbf{OTO(Sum)} & 29.36  & 28.92  & 0.8207  \bigstrut\\
			\cline{2-5}          & \textbf{OTO(Linear)} & 29.34  & 28.93  & \textbf{0.8222 } \bigstrut\\
			\cline{2-5}          & \textbf{OTO} & \textbf{29.36 } & \textbf{28.94 } & \textbf{0.8222 } \bigstrut\\
			\hline
			\multirow{4}[8]{*}{\textbf{QF-20}} & \textbf{JPEG} & 30.12  & 27.50  & 0.8541  \bigstrut\\
			\cline{2-5}          & \textbf{OTO(Sum)} & 31.56  & 31.00  & 0.8767  \bigstrut\\
			\cline{2-5}          & \textbf{OTO(Linear)} & 31.54  & 31.01  & 0.8774  \bigstrut\\
			\cline{2-5}          & \textbf{OTO} & \textbf{31.64 } & \textbf{31.10 } & \textbf{0.8785 } \bigstrut\\
			\hline
			\multirow{4}[8]{*}{\textbf{QF-30}} & \textbf{JPEG} & 31.48  & 28.94  & 0.8844  \bigstrut\\
			\cline{2-5}          & \textbf{OTO(Sum)} & 32.52  & 31.99  & 0.8966  \bigstrut\\
			\cline{2-5}          & \textbf{OTO(Linear)} & 32.93  & 32.28  & 0.9021  \bigstrut\\
			\cline{2-5}          & \textbf{OTO} & \textbf{32.95 } & \textbf{32.33 } & \textbf{0.9022 } \bigstrut\\
			\hline
			\multirow{4}[8]{*}{\textbf{QF-40}} & \textbf{JPEG} & 32.43  & 29.92  & 0.9011  \bigstrut\\
			\cline{2-5}          & \textbf{OTO(Sum)} & 33.46  & 32.91  & 0.9114  \bigstrut\\
			\cline{2-5}          & \textbf{OTO(Linear)} & 33.77  & 33.06  & 0.9139  \bigstrut\\
			\cline{2-5}          & \textbf{OTO} & \textbf{33.85 } & \textbf{33.13 } & \textbf{0.9155 } \bigstrut\\
			\hline
		\end{tabular}%
		\label{tab:Comparisons between OTO and its two variants on Classic 5}%
	\end{table}%

	\begin{table}[htbp]
		\centering
		\caption{ The weight factor $\alpha$ evaluation experiments on LIVE1}
		\begin{tabular}{|c|c|c|c|c|}
			\hline
			\textbf{LIVE1} & \textbf{$\alpha$} & \textbf{PSNR} & \textbf{PSNR-B} & \textbf{SSIM} \bigstrut\\
			\hline
			\multirow{4}[8]{*}{\textbf{QF-20}} & \textbf{0.01} & 31.59  & 31.12  & 0.8947  \bigstrut\\
			\cline{2-5}       & \textbf{0.1} & 31.59  & 31.12  & 0.8949  \bigstrut\\
			\cline{2-5}       & \textbf{1.0} & 31.60  & 31.07  & 0.8941  \bigstrut\\
			\cline{2-5}       & \textbf{0.0651(auto-learned)} & \textbf{31.62 } & \textbf{31.12 } & \textbf{0.8950 } \bigstrut\\
			\hline
			\multirow{4}[8]{*}{\textbf{QF-30}} & \textbf{0.01} & 33.03  & 32.41  & 0.9214  \bigstrut\\
			\cline{2-5}       & \textbf{0.1} & 33.04  & 32.43  & 0.9214  \bigstrut\\
			\cline{2-5}       & \textbf{1.0} & 32.65  & 32.27  & 0.9160  \bigstrut\\
			\cline{2-5}       & \textbf{0.0544(auto-learned)} & \textbf{33.06 } & \textbf{32.45 } & \textbf{0.9218 } \bigstrut\\
			\hline
		\end{tabular}%
		\label{tab:Alpha Experiment}%
	\end{table}%
	
	We further evaluate how the weight factor $\alpha$ affects the final performance. Results are shown in Table \ref{tab:Alpha Experiment}.  $\alpha$ is implemented based on a scale layer of the Caffe platform, which can be updated by the BP algorithm. We can also give a constant $\alpha$ by manually setting the learning rate of this layer to  be 0, so that $\alpha$ keeps unchanged during the training process. Firstly, we revisit  OTO(Linear) and get the learned  $\alpha$, 0.0651 and 0.0544 for QF=20 and 30 respectively. The weight of the small-scale sub-network is 20 times smaller than that of the normal-scale sub-network, indicating that the normal-scale features contain much richer information than small-scale ones. Then, we set $\alpha$ to 0.01, 0.1 and 1.0 (when $\alpha=1.0$, it leads to OTO(Sum)). The results show that when $\alpha$ is set to 0.01 and 0.1, close to the auto-learned value, the performance is slightly worse than OTO(Linear)(learned $\alpha$), but much better than OTO(Sum) ($\alpha =1$), particularly on QF=30. 
	Considering on all cases that OTO(Linear) achieves better results, we can conclude that an auto-learned $\alpha$ is significant for a practical CAR system especially when  a proper   $\alpha$ cannot be given in advance. In addition, OTO(Sum)  means that no difference model is in use, in contrast our OTO with the difference model always achieve a better performance as shown in Table \ref{tab:Comparisons between OTO and its two variants on LIVE1} and Table \ref{tab:Comparisons between OTO and its two variants on Classic 5}, which prove that OTOs benefit from the high frequency information. We visualize the feature maps after the summation model and the difference models in Fig.~\ref{fig:frequency} for a picture from LIVE1. The results show that difference model provides more detailed information(high frequency) than the summation model and it clearly supports our motivation.

	\subsection{On Remote Sensing Image Datasets}\label{On Remote Sensing Image Dataset}
	\begin{table}[htbp]
		\centering
		\caption{Results on Remotely Sensed Dataset}
		\begin{tabular}{|c|c|c|c|c|}
			\hline
			Quality & Evaluation & SPIHT & ARCNN & OTO \bigstrut\\
			\hline
			\multirow{3}[6]{*}{8} & PSNR  & 37.19  & 35.71  & \textbf{39.21 } \bigstrut\\
			\cline{2-5}          & SSIM  & 0.9782  & 0.9775  & \textbf{0.9854 } \bigstrut\\
			\cline{2-5}          & PSNR-B & 31.90  & 33.12  & \textbf{37.95 } \bigstrut\\
			\hline
			\multirow{3}[6]{*}{16} & PSNR  & 33.47  & 32.69  & \textbf{35.23 } \bigstrut\\
			\cline{2-5}          & SSIM  & 0.9523  & 0.9554  & \textbf{0.9652 } \bigstrut\\
			\cline{2-5}          & PSNR-B & 28.80  & 30.61  & \textbf{34.38 } \bigstrut\\
			\hline
			\multirow{3}[6]{*}{32} & PSNR  & 30.47  & 30.39  & \textbf{32.23 } \bigstrut\\
			\cline{2-5}          & SSIM  & 0.9111  & 0.9200  & \textbf{0.9361 } \bigstrut\\
			\cline{2-5}          & PSNR-B & 26.31  & 28.54  & \textbf{31.53 } \bigstrut\\
			\hline
			\multirow{3}[6]{*}{64} & PSNR  & 27.90  & 27.83  & \textbf{29.54 } \bigstrut\\
			\cline{2-5}          & SSIM  & 0.8489  & 0.8547  & \textbf{0.8875 } \bigstrut\\
			\cline{2-5}          & PSNR-B & 24.19  & 26.55  & \textbf{29.20 } \bigstrut\\
			\hline
		\end{tabular}%
		\label{tab:Restoration Rusults on Remotely Sensed Dataset}%
	\end{table}%

\begin{figure}[htbp]
		\begin{center}
			\includegraphics[width=0.95\linewidth]{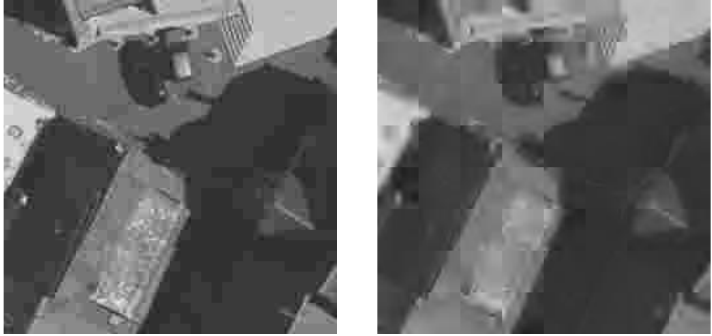}
			\caption{The difference between JPEG and SPIHT compression algorithm. Left: JPEG with block size 8$\times$8, Right: SPIHT with block size 32$\times$32. The blocking artifact caused by SPIHT is  more severe than  by JPEG.}
			\label{fig:88vs3232}
		\end{center}
	\end{figure}

	For JPEG-based compression artifacts reduction methods, their target block size is $8\times8$, but in our SPIHT-based algorithm, blocking artifacts with a larger size like $32\times32$ will occur, which is shown in Fig.~\ref{fig:88vs3232}. Remotely sensed images are quite different from the natural images like BSD500 in terms of  color richness, texture distribution and so on. ARCNN is first designed for restoring natural images. For a fair comparison, we retrain ARCNN on the remote sensed image dataset with the architecture of the network unchanged.  We adopt better training parameters with step-attenuated learning rate compared to its fixed one. The network tends to converge early in the four compression rates, 8, 16, 32 and 64 and then we evaluate it on the remote sensing task. We train and test our OTOs  on remotely sensed image dataset, and the results are shown in Table \ref{tab:Restoration Rusults on Remotely Sensed Dataset}. The parameters in PSNR-B and SSIM algorithms   are modified to evaluate the $32\times32$-sized blocking artifacts.
	
	It is astonishing that in various compression rates,  ARCNN does not increase three scores except for PSNR-B, while OTO successfully suppressed compression artifacts on all measures. In Fig.~\ref{f8} and Fig.~\ref{f9}, the images restored by ARCNN tend to be blurry with blocking artifacts remained.  The failure of ARCNN and the success of the OTO verify that  OTO  are quite effective for remote sensing images restoration, when suffering from larger blocking artifacts problems. However, when compression rate becomes bigger, i.e., 64, the details of the compressed images are almost lost, our OTO fail to restore the edges and structure details of the balcony as shown in Fig.~\ref{f10}. 
	
	\subsection{On LIVE1 and  BSD500 Tests sets}\label{On LIVE1 and  BSD500 Tests set}
	\textbf{LIVE1: }As mentioned above, the proposed OTO outperform ARCNN on the remote sensing image dataset and shows the promising results on restoring SPIHT-based compression artifacts. The following experiments further support that even compared with recently proposed deep learning methods, OTO can still achieve the state-of-the-art results on publicly LIVE1 and BSD500 test sets based on the JPEG compression.
	
	We compare OTO with the most successful deblocking oriented method, SA-DCT, which achieves the state-of-the-art results. Then ARCNN is also included for a complete assessment, using the same metric as before. The results are shown in Table \ref{tab:Restoration Results on LIVE1}. ARCNN does not use data augmentation technique on the training set in the initial conference version, but in its extended journal version  $20\times$ augmentation method is used so as to gain restoration performance improvement. In our experiments, no data augmentation is applied with the aim to accelerate the training process.  Specifically, for the PSNR metric, we achieve an average gain of 0.90 dB compared with SA-DCT and 0.32 dB compared with ARCNN. For the PSNR-B metric, the gains are even larger to 1.38 dB and 0.34 dB respectively. It shows that OTOs are suitable for suppressing compression artifacts for natural images.
	
	\begin{table}[t]
		\small
		\centering
		\caption{ Results on LIVE1}
		\begin{tabular}{|c|c|c|c|c|}
			\hline
			\textcolor[rgb]{ 0,  0,  0}{\textbf{LIVE1}} & \textbf{Algorithm} & \textbf{PSNR} & \textbf{PSNR-B} & \textbf{SSIM} \bigstrut\\
			\hline
			\multirow{4}[8]{*}{\textbf{QF-10}} & \textbf{JPEG} & 27.77  & 25.33  & 0.7905  \bigstrut\\
			\cline{2-5}       & \textbf{SA-DCT} & 28.65  & 28.01  & 0.8093  \bigstrut\\
			\cline{2-5}       & \textbf{AR-CNN} & 29.13  & 28.74  & 0.8232  \bigstrut\\
			\cline{2-5}       & \textbf{OTO} & \textbf{29.28 } & \textbf{28.95 } & \textbf{0.8298 } \bigstrut\\
			\hline
			\multirow{4}[8]{*}{\textbf{QF-20}} & \textbf{JPEG} & 30.07  & 27.57  & 0.8683  \bigstrut\\
			\cline{2-5}       & \textbf{SA-DCT} & 30.81  & 29.82  & 0.8781  \bigstrut\\
			\cline{2-5}       & \textbf{AR-CNN} & 31.40  & 30.69  & 0.8886  \bigstrut\\
			\cline{2-5}       & \textbf{OTO} & \textbf{31.67 } & \textbf{31.17 } & \textbf{0.8954 } \bigstrut\\
			\hline
			\multirow{4}[8]{*}{\textbf{QF-30}} & \textbf{JPEG} & 31.41  & 28.92  & 0.9000  \bigstrut\\
			\cline{2-5}       & \textbf{SA-DCT} & 32.08  & 30.92  & 0.9078  \bigstrut\\
			\cline{2-5}       & \textbf{AR-CNN} & 32.69  & 32.15  & 0.9166  \bigstrut\\
			\cline{2-5}       & \textbf{OTO} & \textbf{33.08 } & \textbf{32.48 } & \textbf{0.9218 } \bigstrut\\
			\hline
			\multirow{4}[8]{*}{\textbf{QF-40}} & \textbf{JPEG} & 32.35  & 29.96  & 0.9173  \bigstrut\\
			\cline{2-5}       & \textbf{SA-DCT} & 32.99  & 31.79  & 0.9240  \bigstrut\\
			\cline{2-5}       & \textbf{AR-CNN} & 33.63  & 33.12  & 0.9306  \bigstrut\\
			\cline{2-5}       & \textbf{OTO} & \textbf{34.10 } & \textbf{33.48 } & \textbf{0.9362 } \bigstrut\\
			\hline
		\end{tabular}%
		\label{tab:Restoration Results on LIVE1}%
	\end{table}%
	
	\begin{table}[htbp]
		\small
		\centering
		\caption{ Results on BSD500 Test Set}
		\begin{tabular}{|c|c|c|c|c|c|c|c|}
			\hline
			Quality & Evaluation & JPEG  & DSC   & DDCN(-DCT) & TNRD  & ARCNN & OTO \bigstrut\\
			\hline
			\multirow{3}[6]{*}{10} & PSNR  & 27.8  & 28.79 & 29.26 & 29.16 & 29.10  & \textbf{29.31 } \bigstrut\\
			\cline{2-8}          & SSIM  & 0.7875 & 0.8124 & 0.8267 & 0.8225 & 0.8198 & \textbf{0.8278} \bigstrut\\
			\cline{2-8}          & PSNR-B & 25.1  & 28.45 & 28.89 & 28.81 & 28.73 & \textbf{28.92 } \bigstrut\\
			\hline
			\multirow{3}[6]{*}{20} & PSNR  & 30.05 & 30.97 & 31.55 & 31.41 & 31.28 & \textbf{31.64 } \bigstrut\\
			\cline{2-8}          & SSIM  & 0.8671 & 0.8804 & 0.8923 & 0.8889 & 0.8854 & \textbf{0.8943} \bigstrut\\
			\cline{2-8}          & PSNR-B & 27.22 & 30.57 & 30.84 & 30.83 & 30.55 & \textbf{30.95 } \bigstrut\\
			\hline
			\multirow{3}[6]{*}{30} & PSNR  & 31.37 & 32.29 & 32.92 & 32.77 & 32.67 & \textbf{33.03 } \bigstrut\\
			\cline{2-8}          & SSIM  & 0.8994 & 0.9093 & 0.9193 & 0.9166 & 0.9152 & \textbf{0.9211} \bigstrut\\
			\cline{2-8}          & PSNR-B & 28.53 & 31.84 & 32.01 & 31.99 & 31.94 & \textbf{32.17 } \bigstrut\\
			\hline
			\multirow{3}[6]{*}{40} & PSNR  & 32.3  & 33.23 & 33.87 & 33.73 & 33.55 & \textbf{34.00 } \bigstrut\\
			\cline{2-8}          & SSIM  & 0.9171 & 0.9253 & 0.9336 & 0.9316 & 0.9296 & \textbf{0.9357} \bigstrut\\
			\cline{2-8}          & PSNR-B & 29.49 & 32.71 & 32.86 & 32.79 & 32.78 & \textbf{32.98 } \bigstrut\\
			\hline
		\end{tabular}%
		\label{tab:Restoration Results on BSD500 Test Set}%
	\end{table}%

	\textbf{BSD500: }We compare OTO with the traditional approaches like DSC and also convolutional deep learning based approaches, such as ARCNN and Trainable Nonlinear Reaction Diffusion (TNRD)  \cite{chen2015learning}. In DDCN, the DCT-Domain branch took advantage of JPEG-based prior so it is unfair for OTO only using pixel-domain information. Guo et al. propose a variant of DDCN by removing the DCT-domain branch so that no extra prior is utilized, which is alternatively used in the comparison.  The comparative results are shown in Table \ref{tab:Restoration Results on BSD500 Test Set} with four quality factors from 10 to 40. OTO outperforms all the other algorithms in terms of three metrics, which indicates that OTO has a competent restoration ability. More specifically, OTO obtains about 0.7 dB and 0.4 dB gains compared with DSC on the PSNR and PSNR-B respectively. ARCNN is beaten by 0.35 dB on the PSNR and 0.26 dB on the PSNR-B, which is consistent with the results on LIVE1.
	\begin{figure*}[h!]
		\begin{center}
			\includegraphics[width=0.95\linewidth]{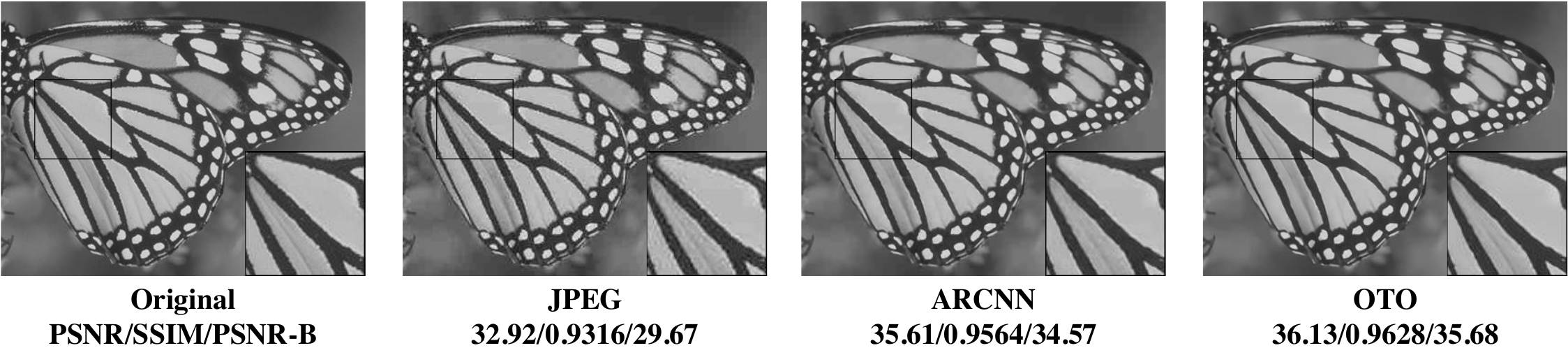}
			\caption{Qualitative comparison of OTO and ARCNN by JPEG with Quality Factor=20 where ringing effects is carefully handled after being restored by OTO network.}
			\label{f6}
		\end{center}
	\end{figure*}
	
	\begin{figure*}[h!]
		\begin{center}
			\includegraphics[width=0.95\linewidth]{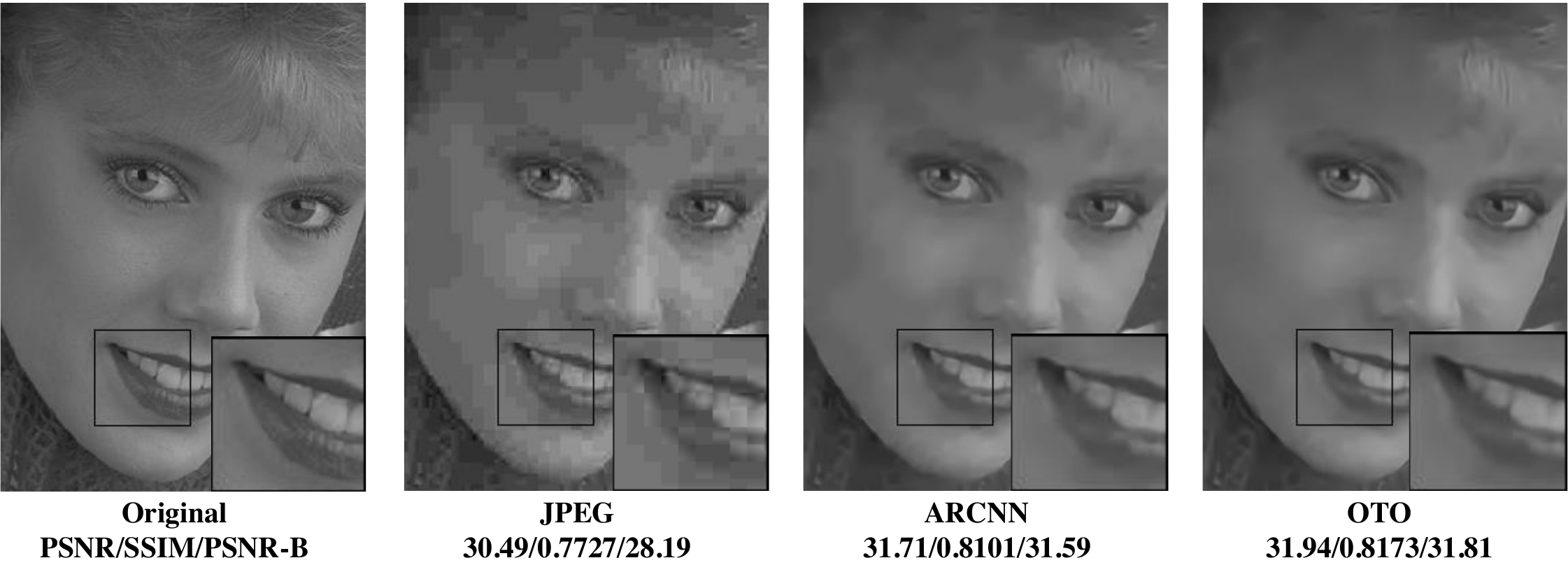}
			\caption{Qualitative comparison of OTO and ARCNN by JPEG with Quality Factor=10, where severe block artifacts are removed and the edges are sharp again.}
			\label{f7}
		\end{center}
	\end{figure*}
	
	\begin{figure*}[h!]
		\begin{center}
			\includegraphics[width=0.95\linewidth]{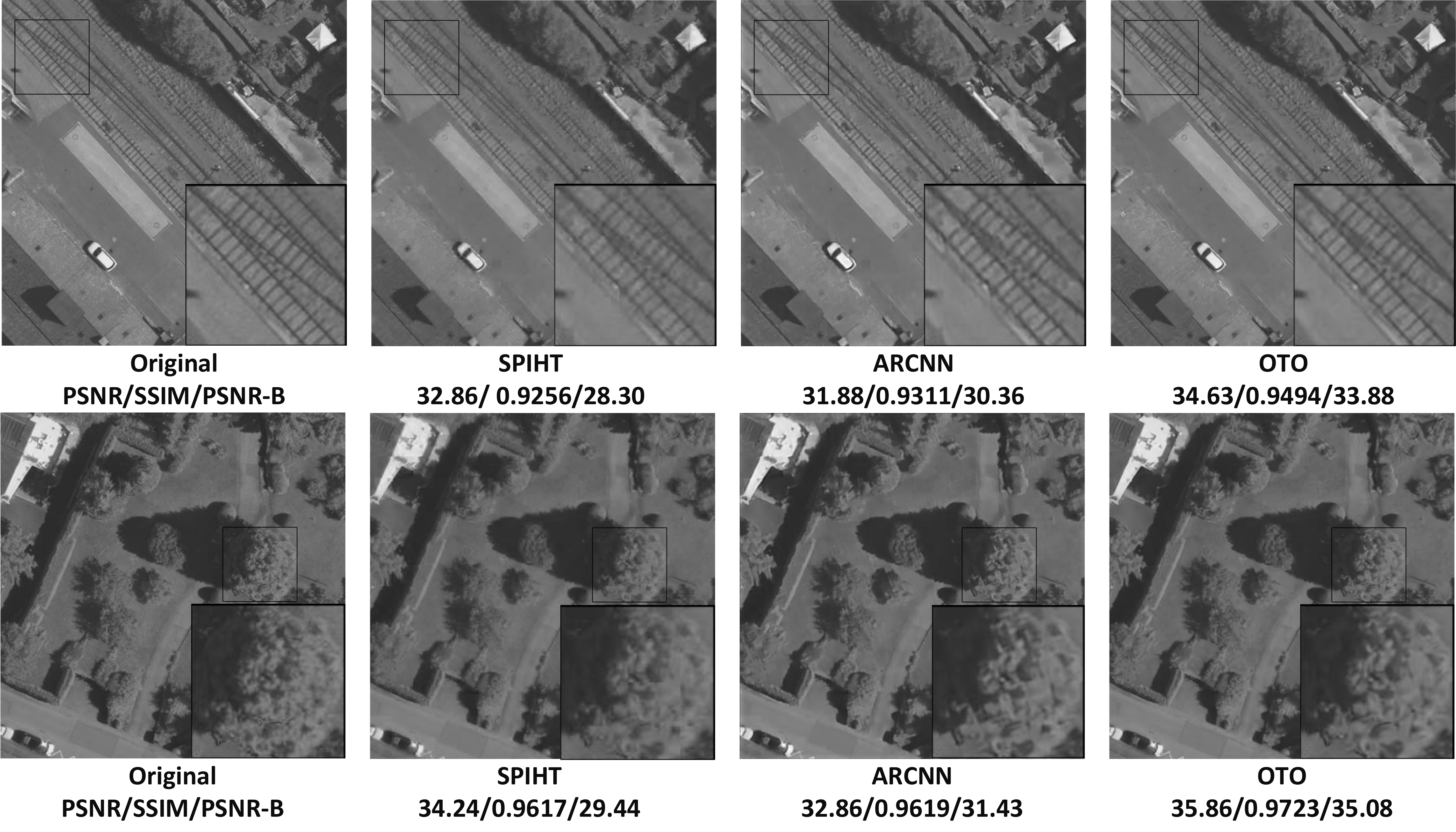}
			\caption{Qualitative comparison of OTO and ARCNN by SPIHT with Compression Rate=16.}
			\label{f8}
		\end{center}
	\end{figure*}
	
	\begin{figure*}[h!]
		\begin{center}
			\includegraphics[width=0.95\linewidth]{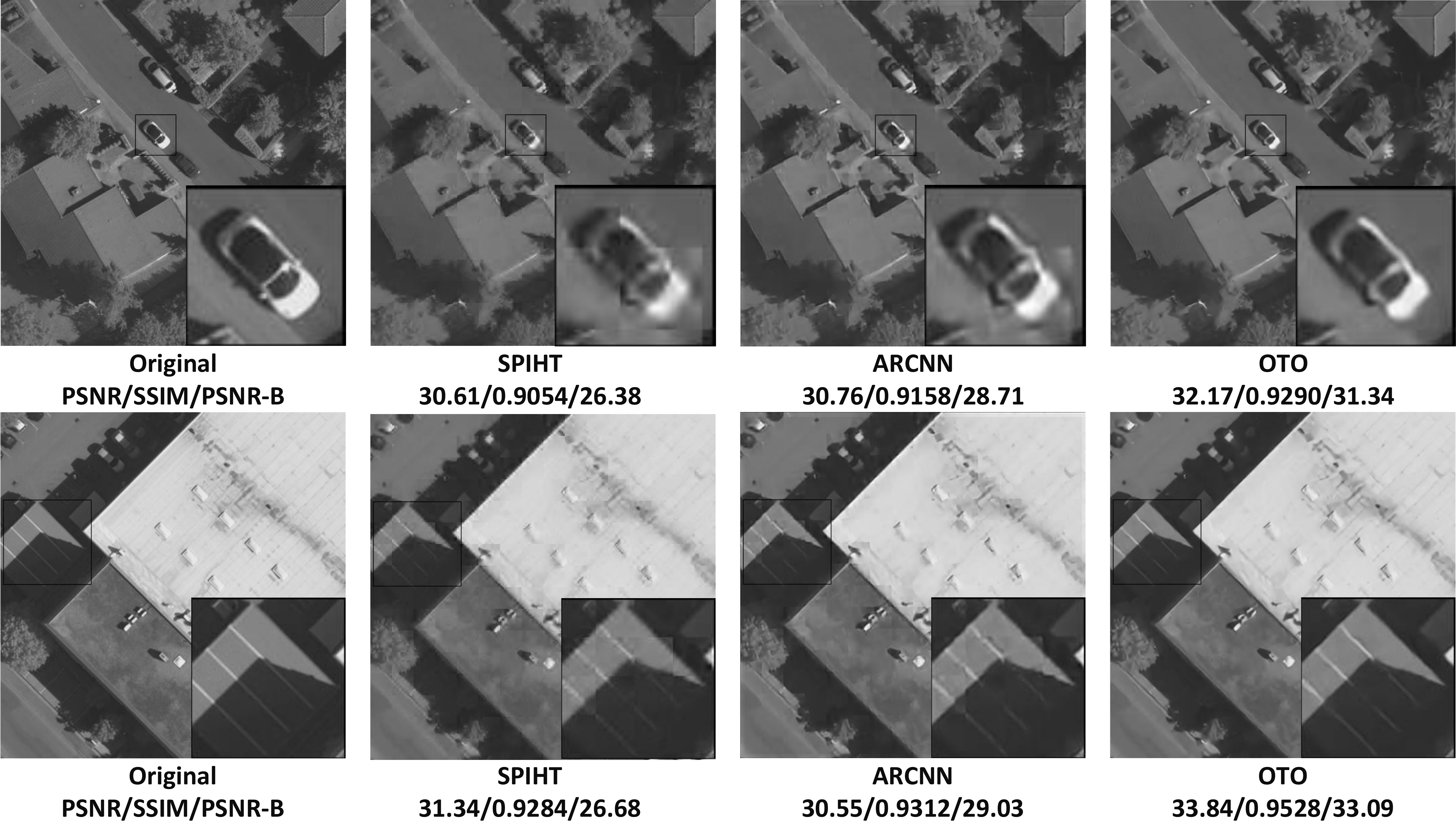}
			\caption{Qualitative comparison of OTO and ARCNN by SPIHT with Compression Rate=32.}
			\label{f9}
		\end{center}
	\end{figure*}
	
	\begin{figure*}[h!]
		\begin{center}
			\includegraphics[width=0.95\linewidth]{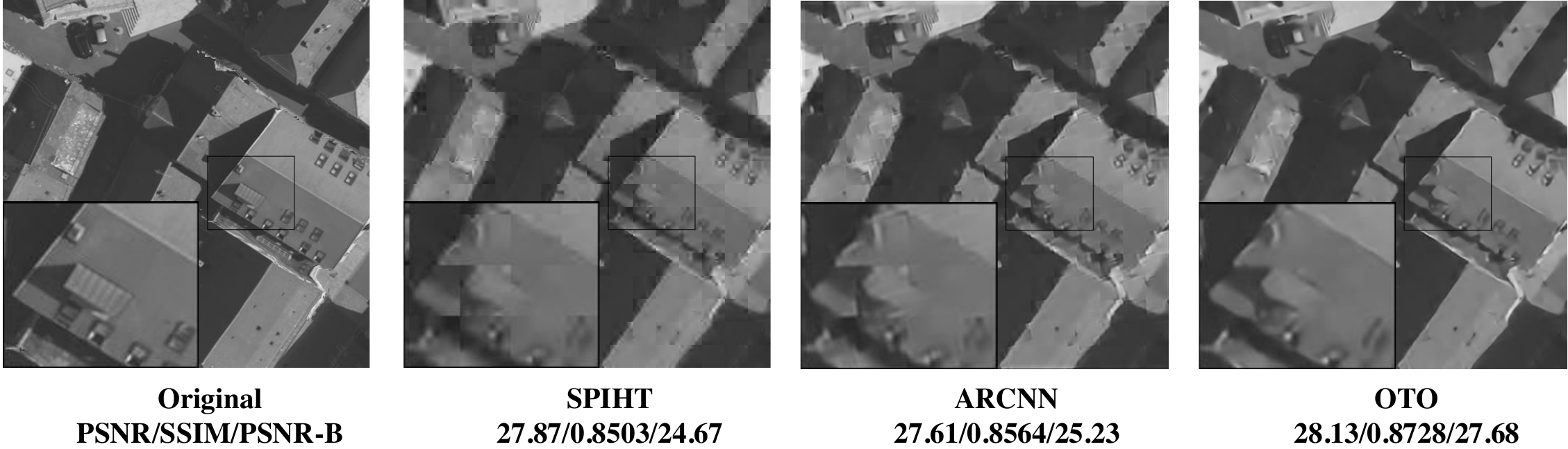}
			\caption{Qualitative comparison of OTO and ARCNN by SPIHT with Compression Rate=64.}
			\label{f10}
		\end{center}
	\end{figure*}
	
	\section{Conclusion and future work}\label{Conclusion and future work}
	
	The CAR problem is a challenge in the field of remote sensing. In this paper, we have developed a new and general framework to combine different models based on a nonlinear method to effectively deal with complicated compression artifacts, i.e., big blocking effect in the compression. Based on the Taylor expansion, we lead to two simple OTO variants, which provide a more profound investigation into our method and  pose a new direction to solve the artifact reduction problem. Extensive experiments are conducted to validate the performance of OTO and new state-of-the-art results are obtained. In the future work, we will deploy more complicated networks in our framework to gain better performance.
	
	%
	%
	\section*{References}
	\bibliography{bibliography}
	\section*{Biography}
	\paragraph{Baochang Zhang}
	received the B.S., M.S. and Ph.D. degrees in Computer Science from Harbin Institute of the Technology, Harbin, China, in 1999, 2001, and 2006, respectively. From 2006 to 2008, he was a research fellow with the Chinese University of Hong Kong, Hong Kong, and with Griffith University, Brisban, Australia. Currently, he is an associate professor with the Science and Technology on Aircraft Control Laboratory, School of Automation Science and Electrical Engineering, Beihang University, Beijing, China. He was supported by the Program for New Century Excellent Talents in University of Ministry of Education of China. His current research interests include pattern recognition, machine learning, face recognition, and wavelets.
	
	\paragraph{Jiaxin Gu}
	received the B.S. degree in School of Automation Science and Electrical Engineering of Beihang University in 2017. He is pursuing his Master degree in the same shool of Beihang University and his current research interests include image restoration, object detection and deep learning.
	
	\paragraph{Chen Chen}
	received the B.E. degree in automation
	from the Beijing Forestry University, Beijing, China,
	in 2009, and the M.S. degree in electrical engineering
	from the Mississippi State University, Starkville, MS,
	USA, in 2012, and the Ph.D. degree in the Department
	of Electrical Engineering, University of Texas
	at Dallas, Richardson, TX, USA, in 2016.
	He is currently a Post-Doc in the Center for Research
	in Computer Vision, University of Central
	Florida, Orlando, FL, USA. He has published more
	than 50 papers in refereed journals and conferences
	in these areas. His research interests include compressed sensing, signal and
	image processing, pattern recognition, and computer vision.
	
	\paragraph{Jungong Han} is currently a tenured faculty member with the School of Computing and Communications at Lancaster University, UK. His research interestes include video analysis, computer vision and artificial intelligence.
	
	\paragraph{Xiangbo Su} received his B.E. degree in Automation Science from Beihang University, Beijing, China, in 2015. He is currently a M.S. student at the School of Automation Science and
	Electrical Engineering, Beihang University. His current research interests include machine learning and deep learning in general, with computer vision applications in object tracking, recognition and image restoration.
	
	\paragraph{Xianbin Cao}
	received the B.Eng and M.Eng degrees in computer applications and information science from Anhui University, Hefei, China, in 1990 and 1993, respectively, and the Ph.D. degree in information science from the University of Science and Technology of China, Beijing, in 1996. He is currently a Professor with the School of Electronic and Information Engineering, Beihang University, Beijing, China. He is also the Director of the Laboratory of Intelligent Transportation System. His current research interests include intelligent transportation systems, airspace transportation management, and intelligent computation.
	
	\paragraph{Jianzhuang Liu}
	received the Ph.D. degree in computer vision from The Chinese University of Hong Kong, Hong Kong, in 1997. He was a Research Fellow with Nanyang Technological University, Singapore, from 1998 to 2000. From 2000 to 2012, he was a Post-Doctoral Fellow, an Assistant Professor, and an Adjunct Associate Professor with The Chinese University of Hong Kong. He was a Professor in 2011 with the University of Chinese Academy of Sciences. He is currently a Principal Researcher with Huawei Technologies Company, Ltd., Shenzhen. He has authored over 100 papers, most of which are in prestigious journals and conferences in computer science. His research interests include computer vision, image processing, machine learning, multimedia, and graphics.
	
\end{document}